\documentclass[runningheads]{llncs}
\usepackage[T1]{fontenc}
\usepackage{graphicx,verbatim}
\usepackage{booktabs}
\usepackage{subfig}

\usepackage[table,xcdraw]{xcolor}
\usepackage{hyperref}
\usepackage{color}

\usepackage{amsmath}
\DeclareMathOperator*{\argmin}{arg\,min}
\usepackage{amssymb}
\usepackage{cleveref}

\begin{document}
\title{BridgeSplat: Bidirectionally Coupled CT and Non-Rigid Gaussian Splatting for Deformable Intraoperative Surgical Navigation}
\author{
Maximilian Fehrentz\inst{1, 2} \and
Alexander Winkler\inst{2} \and
Thomas Heiliger\inst{2} \and
Nazim~Haouchine\inst{3} \and
Christian Heiliger\inst{2} \and
Nassir Navab\inst{1}
}
%
\authorrunning{M. Fehrentz et al.}
%
\institute{
Computer Aided Medical Procedures, TU Munich, Munich, Germany\and
Department of General, Visceral, and Transplantation Surgery, Hospital of the LMU Munich, Ludwig-Maximilians-Universität (LMU), Munich, Germany\and
Harvard Medical School, Brigham and Women's Hospital, Boston, MA, USA
}


\maketitle              
\begin{abstract}
We introduce BridgeSplat, a novel approach for deformable surgical navigation that couples intraoperative 3D reconstruction with preoperative CT data to bridge the gap between surgical video and volumetric patient data. Our method rigs 3D Gaussians to a CT mesh, enabling joint optimization of Gaussian parameters and mesh deformation through photometric supervision. By parametrizing each Gaussian relative to its parent mesh triangle, we enforce alignment between Gaussians and mesh and obtain deformations that can be propagated back to update the CT. We demonstrate BridgeSplat's effectiveness on visceral pig surgeries and synthetic data of a human liver under simulation, showing sensible deformations of the preoperative CT on monocular RGB data. Code, data, and additional resources can be found at \href{https://maxfehrentz.github.io/ct-informed-splatting/}{https://maxfehrentz.github.io/ct-informed-splatting/}.


\end{abstract}

\section{Introduction}
Laparoscopic surgery has shown the potential to lower complication rates, shorten hospital stays, and be cost-effective while maintaining R0 resection rates and showing no adverse effects on 90-day mortality \cite{fretland_laparoscopic_2018}. However, it also comes with challenges, as tactile feedback is limited, spatial understanding is demanding, handling is challenging, and learning curves are flat. To mitigate those issues, Augmented Reality (AR) has been proposed to overlay relevant information from preoperative scans with little interruption to the clinical workflow. The potential for AR-guided (minimally invasive) interventions in visceral surgery has been demonstrated for the liver 
\cite{bertrand_case_2020}
, kidney 
\cite{piana_automatic_2024}
, pancreas 
\cite{wu_augmented_2023}
, and spleen \cite{tao_application_2021}. Although the results are promising, those methods either stop at a rigid registration or require interruption of the surgical workflow for surgeon interaction (e.g., outlining contours). Continuous non-rigid registration is still an open challenge.


Thinking from first principles, we would like a system that can reconstruct the intraoperative scene in real time while also estimating the camera pose. Therefore, much effort has been directed at non-rigid SLAM for surgery 
\cite{song_mis-slam_2018,rodriguez_nr-slam_2023,tang_real-time_2024}. 
Despite all the progress on SLAM, as concluded in \cite{shu_seamless_2024}, tracking of the camera is still desirable, if not necessary, to move towards clinically feasible systems.

For posed cameras in surgical scenes, methods capable of 4D reconstruction for dynamic photorealistic novel view synthesis have emerged. Those methods are grounded in NeRF \cite{endonerf} and 4D Gaussian Splatting (4DGS) \cite{zhu_endogs_2024,liu_lgs_2024,li_endosparse_2024,yang_deform3dgs_2024,linguraru_endo-4dgs_2024}. Prior work has been done based on both NeRF and 4DGS to go beyond mere reconstruction and towards intelligence, tracking the surgical scene \cite{gerats2024neural,Hay_Online_MICCAI2024}. However, NeRF does not have an explicit representation, and 4DGS is usually not bound to a surface. Attempting a monocular single-view reconstruction commonly leads to artifacts like floaters in the view frustum that allow overfitting on the individual observations but do not guarantee a meaningful 3D reconstruction of tissue over time, despite high image reconstruction quality.

In \cite{wang_video-based_2024}, the authors propose to extract a preoperative 3D model, equip it with a biomechanical model, and perform an initial rigid registration based on an intraoperative point cloud acquired with stereo depth. To allow for a non-rigid registration, they track the registered structure and deform the biomechanical model, using optical flow. Also building on biomechanical models, the authors in \cite{docea2022laparoscopic} propose a CNN for non-rigid registration.

Although some of the introduced methods leverage \emph{selected} 3D information from preoperative scans, none of them is \emph{directly coupled} to the preoperative scan, using the entire 3D prior for guiding the reconstruction and enabling direct deformation propagation to the CT.


\subsubsection{Contribution}
We deform a preoperative CT from an intraoperative \emph{monocular} RGB video. By rigging Gaussians to a registered CT mesh, we constrain the ill-posed monocular reconstruction problem through a canonical space prior, resolving the depth ambiguity. This reframes the problem of non-rigid registration from a non-rigid reconstruction to a non-rigid tracking problem with the preoperative CT as template. Unlike existing 4D Gaussian Splatting methods, our approach is bound to patient-specific anatomy, and its deformations are directly coupled to the CT.

\section{Method}

\begin{figure}
    \centering
    \includegraphics[width=1\linewidth]{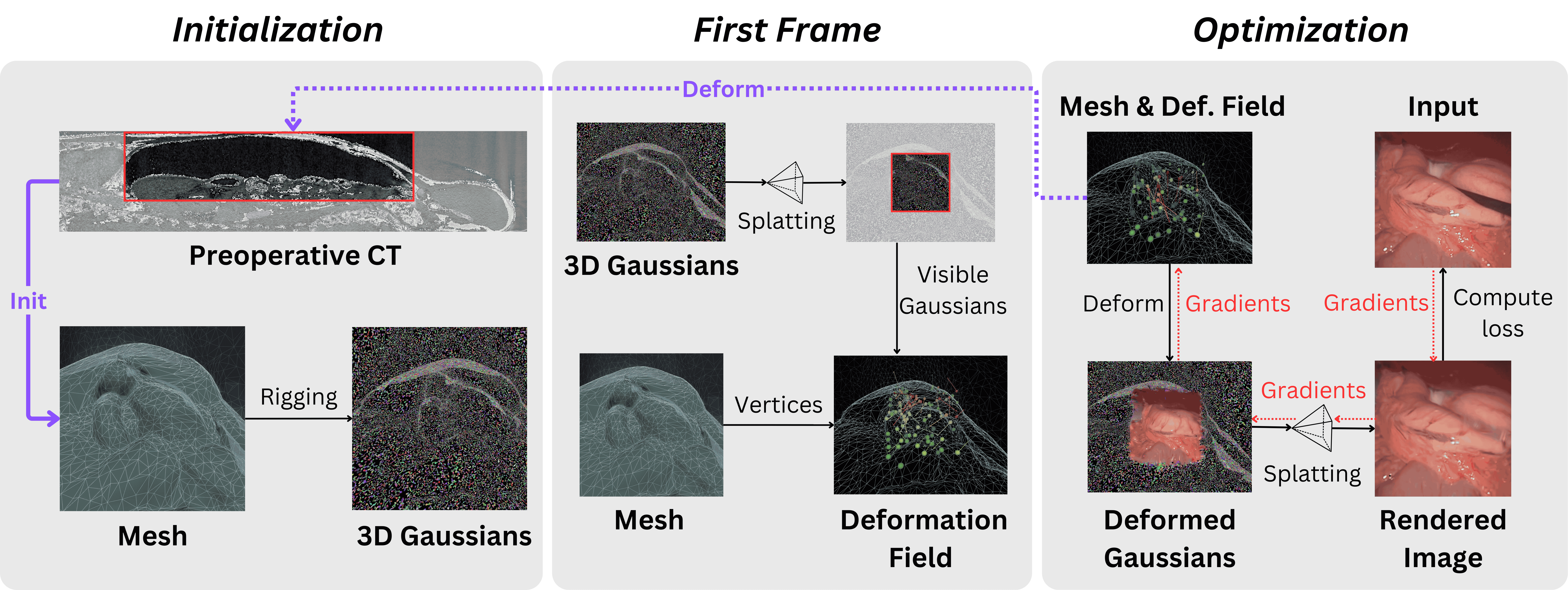}
    \caption{We initialize 3D Gaussians by rigging them onto the registered mesh, extracted from the preoperative CT. Given the first camera pose, we identify the visible Gaussians, derive the visible vertices, and subsample them to initialize a sparse deformation field. For all consecutive frames, we optimize for the mesh vertices, the deformation vectors, and the appearance properties of the 3D Gaussians. Deformations are propagated to the CT.} 
    \label{fig:method-overview}
\end{figure}

\subsection{Problem Formulation}
We start with a preoperative mesh \(\mathbf{M}\) of the abdominal cavity, obtained from the preoperative CT. At the beginning of surgery, a rigid registration \(\mathbf{T}_{CT}\) is performed to align the CT (and thus \(\mathbf{M}\)) with the patient. We then introduce a set of 3D Gaussians, which are rigged onto the registered mesh \(\mathbf{T}_{CT}\mathbf{M}\). The 3D Gaussians have learnable parameters \(\boldsymbol{\Theta}_t\ = \{\boldsymbol{\mu}_t,\;\boldsymbol{\alpha}_t,\;\mathbf{\Sigma}_t,\;\mathbf{c}_t\}\): means \(\boldsymbol{\mu}_t\), opacities \(\boldsymbol{\alpha}_t\), covariance matrices \(\mathbf{\Sigma}_t\), and spherical harmonic coefficients \(\mathbf{c}_t\). Additionally, a deformation field \(\theta_t\) is defined to capture mesh deformations over time.

During surgery, a tracked laparoscope provides images \(\mathbf{I}_t\) from known camera poses \(\mathbf{P}_t\). Our main objective is to minimize a photometric energy term \(\mathbf{E}_\mathrm{photo}\) between the observed image \(\mathbf{I}_t\) and a rendered image \(\hat{\mathbf{I}}_t\) via 3D Gaussian Splatting of the \emph{deformed} mesh and its rigged 3D Gaussians. Formally, we seek to solve:

\begin{equation}
\argmin_{\theta_t,\;\boldsymbol{\Theta}_t}\,\mathbf{E}_{\mathrm{photo}}
\;=\;
\argmin_{\theta_t,\;\boldsymbol{\Theta}_t}
\Bigl\|\,
   \mathbf{I}_t 
   \;-\;
   \hat{\mathbf{I}}_t\!\bigl(
      \mathbf{P}_t,\;
      \theta_t(\mathbf{T}_{CT}\mathbf{M}),\;
      \boldsymbol{\Theta}_t
   \bigr)
\Bigr\|.
\label{eq:frame_optimization}
\end{equation}

where \(\theta_t\) deforms the registered mesh \(\mathbf{T}_{CT}\mathbf{M}\).

\subsection{Coupling 3D Gaussians and Mesh}
Unlike conventional 3D Gaussian Splatting methods that treat the means \(\boldsymbol{\mu}_t\) and covariances \(\mathbf{\Sigma}_t\) as unconstrained parameters, we tightly anchor our Gaussians to the mesh, inspired by GaussianAvatars \cite{qian_gaussianavatars_2024}. Specifically, each triangle \(f^i\) of the deformed mesh \(\theta_t(\mathbf{T}_{CT}\mathbf{M})\) holds one or more Gaussians whose means are expressed via optimizable barycentric coordinates:
\begin{equation}
    \boldsymbol{\mu}_t^i
    \;=\;
    \sum_{k=1}^{3}
    \mathbf{b}_{t}^{i,k} \,\mathbf{v}_t^{i,k},
    \quad
    \text{with}
    \quad
    \sum_{k=1}^{3} \mathbf{b}_{t}^{i,k} \;=\; 1,
    \quad
    \mathbf{b}_{t}^{i,k} \;\ge\; 0,
\label{eq:barycentric_mu}
\end{equation}
where \(\mathbf{v}_t^{i,k} = \theta_t\bigl(\mathbf{v}_{\mathrm{init}}^{i,k}\bigr)\) are the vertices of triangle \(f^i\) after deformation, and \(\mathbf{b}_t^i = (b_{t}^{i,1}, b_{t}^{i,2}, b_{t}^{i,3})\) are the barycentric coordinates for the \(i\)-th Gaussian. By enforcing nonnegative \(\mathbf{b}_{t}^{i,k}\) that sum to 1, the mean \(\boldsymbol{\mu}_t^i\) cannot leave the parent face. This is in contrast to GaussianAvatars, where Gaussians are allowed to wander off the surface for enhanced visual fidelity. In our case, however, we try to estimate mesh deformations from single-view monocular RGB input. Therefore, we have to ensure that what we render is closely related to what the mesh actually represents, otherwise, Gaussians can satisfy the photometric energy by roaming and stretching freely beyond their parent triangles without the mesh having to deform.

For \(\mathbf{\Sigma}_t^i\), we decompose the covariance into a rotation (orientation) and an anisotropic scale. In particular, we fix the Gaussian’s orientation \(\mathbf{q}_t^i\) to align with the triangle’s normal, thus keeping the Gaussian aligned with the surface of its parent triangle. The corresponding scale vector \(\mathbf{s}_t^i\) has only \emph{two} freely optimizable components, the component in normal direction is fixed to a small constant. We further cap the remaining scales to keep them within a factor of the initial scale, which is computed from a mesh-dependent heuristic. This prevents the Gaussians from stretching significantly beyond their parent triangles.


\subsection{Deformation Field}
As shown in \Cref{eq:barycentric_mu}, the means of the Gaussians are parametrized w.r.t to $\mathbf{M}$ by the barycentric coordinates and a sparse deformation field $\theta_t$ for each timestep $t$ that is acting on the mesh. We mostly follow the deformation field parametrization as proposed in \cite{Hay_Online_MICCAI2024}. However, instead of defining control points based on anchor 3D Gaussians, we choose control vertices. For the first frame, we perform a single render to check for visible 3D Gaussians. We can trace those back to their parent faces and obtain all visible vertices. We then choose a random subset of anchor vertices ${\mathbf{a}_1, ..., \mathbf{a}_k} \in \mathbf{M}$, where each anchor vertex has a learnable deformation ${\delta_1, ..., \delta_k} \in \mathbb{R}^3$. It provides continuous deformations through interpolation as described in \cite{Hay_Online_MICCAI2024}. Note that choosing initially visible vertices as anchors comes with the downside of modeling deformations only in a fixed field of view and its adjacency. For our short sequences, this is not an issue. For more extensive camera sweeps, the method can be easily extended to continuously check for visibility and add control vertices as the camera moves.

\subsection{Non-rigid Regularization}
To regularize the deformations, we employ an As-Rigid-As-Possible (ARAP) term \(\mathbf{E}_{\mathrm{ARAP}}\)~\cite{arap}. 
Note that this would be infeasible in a mesh-free approach, since ARAP inherently uses mesh topology 
and cannot be applied directly to unconstrained Gaussian Splatting.
Following \cite{Hay_Online_MICCAI2024}, we additionally penalize relative position changes of neighboring vertices and deformations in currently invisible areas.

For clarity, we omit the time dependency in the notation below. ARAP operates on a set of deformed vertices 
\(\{\mathbf{v}'_1, \dots, \mathbf{v}'_n\} \subset \mathbb{R}^3\)\ of a deformed mesh \(\mathbf{M}_\mathrm{deform}\). 
For each vertex \(\mathbf{v}'_i \in \mathbb{R}^3\), ARAP measures the discrepancy between the edges in $\mathbf{M}_\mathrm{deform}$ 
and the corresponding edges in the original mesh \(\mathbf{M}\) under an estimated rotation 
\(\mathbf{R}_i \in \mathrm{SO}(3)\) for each vertex. This encourages local transformations to remain as rigid as possible. In its standard formulation, ARAP is usually driven by a small number of known control points, 
and the minimization of \(\mathbf{E}_{\mathrm{ARAP}}\) proceeds via a flip-flop scheme: 
first, one estimates the per-vertex rotations 
\(\{\mathbf{R}_1, \dots, \mathbf{R}_n\}\subset \mathrm{SO}(3)\) using SVD, 
then one updates the vertex positions 
\(\{\mathbf{v}'_1, \dots, \mathbf{v}'_n\} \subset \mathbb{R}^3\) by directly solving a linear system. 
In our approach, we first set \(\{\mathbf{v}'_1, \dots, \mathbf{v}'_n\} = \theta(\mathbf{M})\). Then, we simply compute the rotations \(\{\mathbf{R}_1, \dots, \mathbf{R}_n\}\) via SVD as proposed in the original ARAP. However, we do not solve for the optimal vertex positions directly. Instead, we compute $\nabla_{\mathbf{v}'_i}\,\mathbf{E}_{\mathrm{ARAP}}$ as derived in the original ARAP paper and use this term in our gradient-based optimization.
\section{Experiments and Results}
We evaluate our method on two datasets. Given that intraoperative data does not come with ground truth deformations, we simulate tool-tissue interaction sequences to evaluate our method quantitatively. To demonstrate its ability to operate on clinical data and deform a CT, we use a second dataset from visceral pig surgeries. All experimental protocols were approved by the local Ethical Committee on Animal Experimentation. Since no comparable methods are available, we cannot compare against a state-of-the-art. Whereas other 4DGS methods are primarily concerned with high-quality novel view synthesis, we focus on the deformation of the mesh and CT and therefore do not compare image reconstruction metrics. Results are best viewed in video format, we refer the reader to the supplementary material.

\subsubsection{Quantitative Evaluation on Simulated Data}
\begin{figure}
    \centering
    \subfloat{\includegraphics[width=0.195\linewidth]{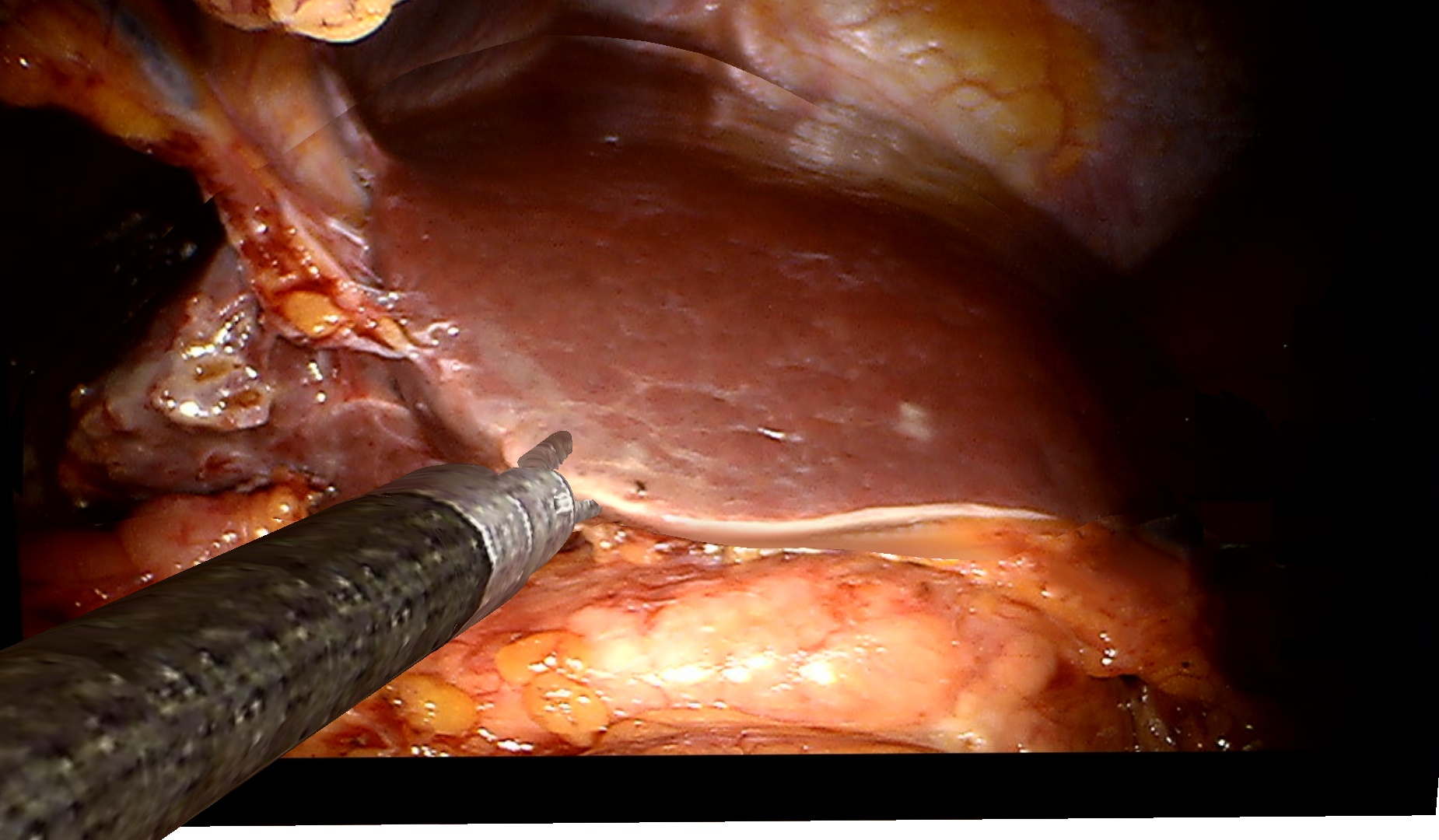}}
    \hfill
    \subfloat{\includegraphics[width=0.195\linewidth]{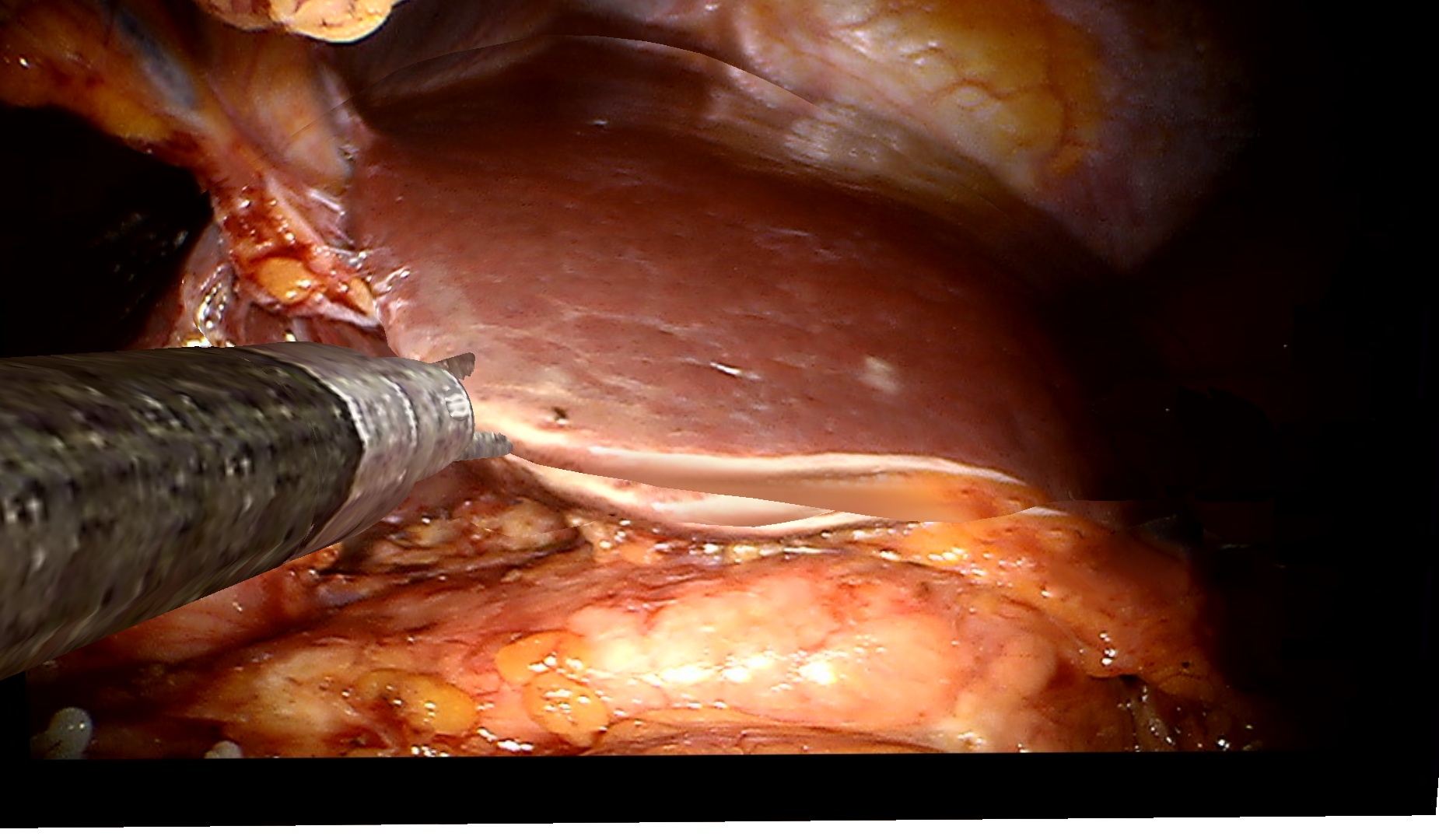}}
    \hfill
    \subfloat{\includegraphics[width=0.195\linewidth]{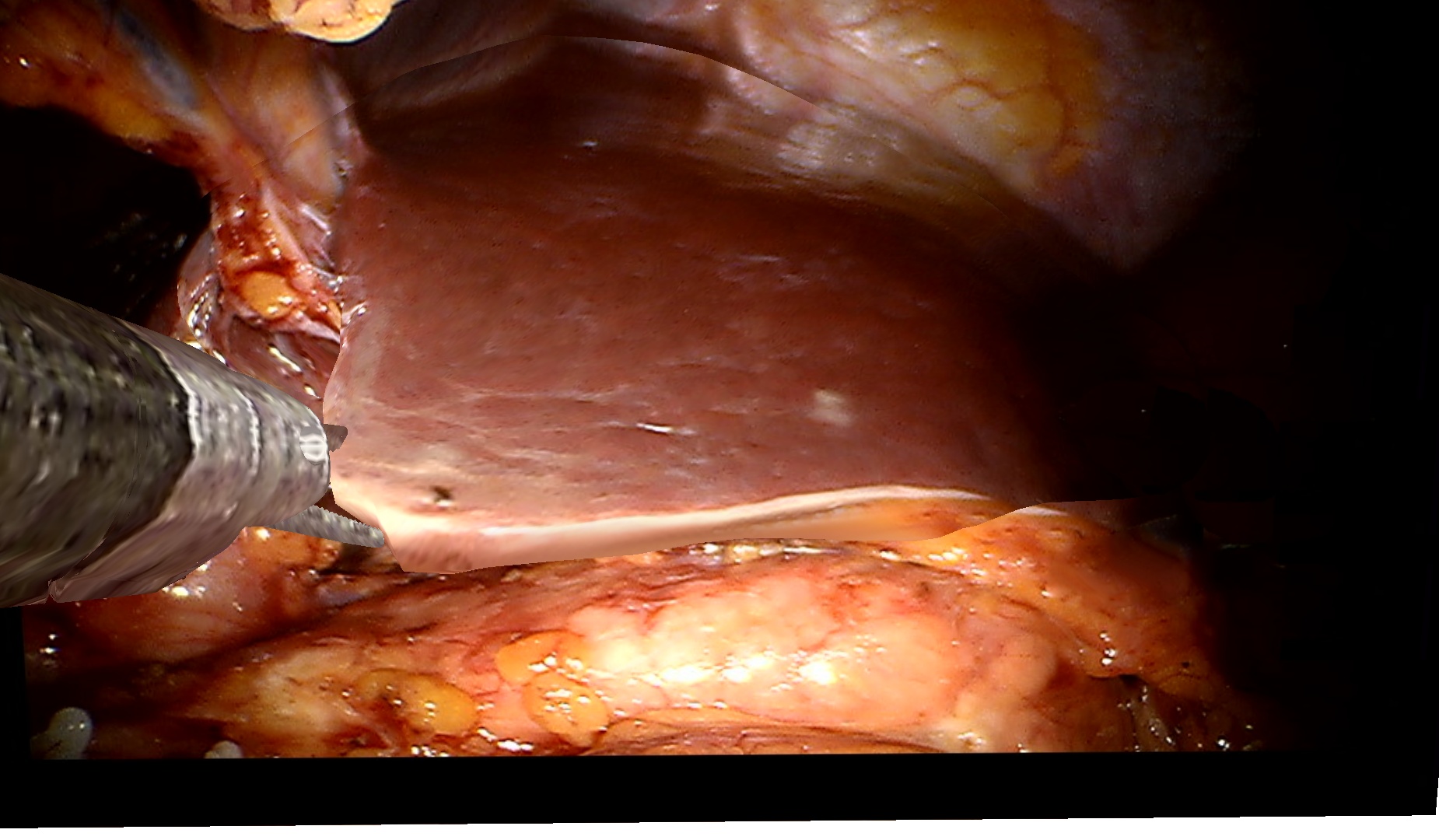}}
    \hfill
    \subfloat{\includegraphics[width=0.195\linewidth]{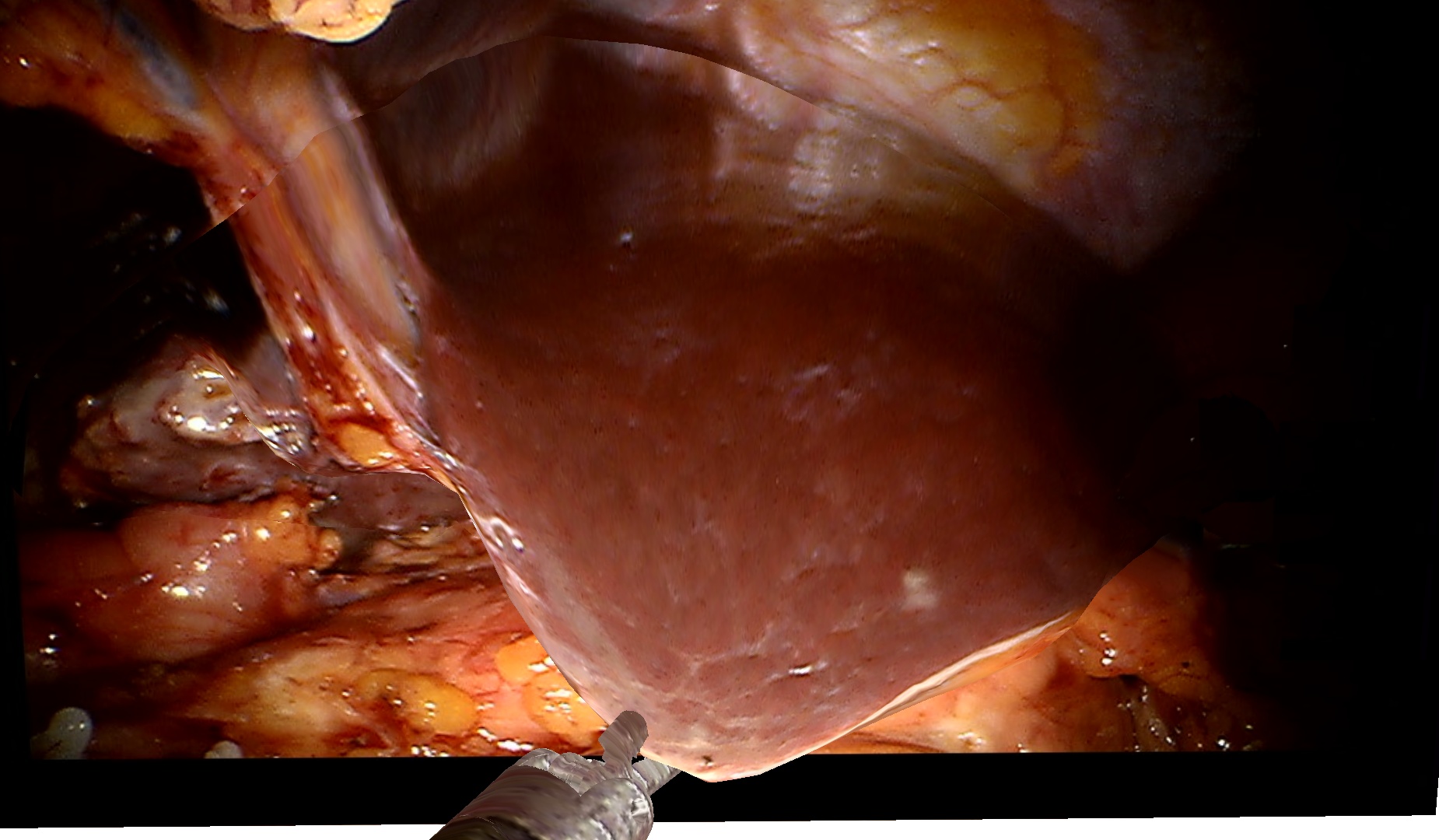}}
    \hfill
    \subfloat{\includegraphics[width=0.195\linewidth]{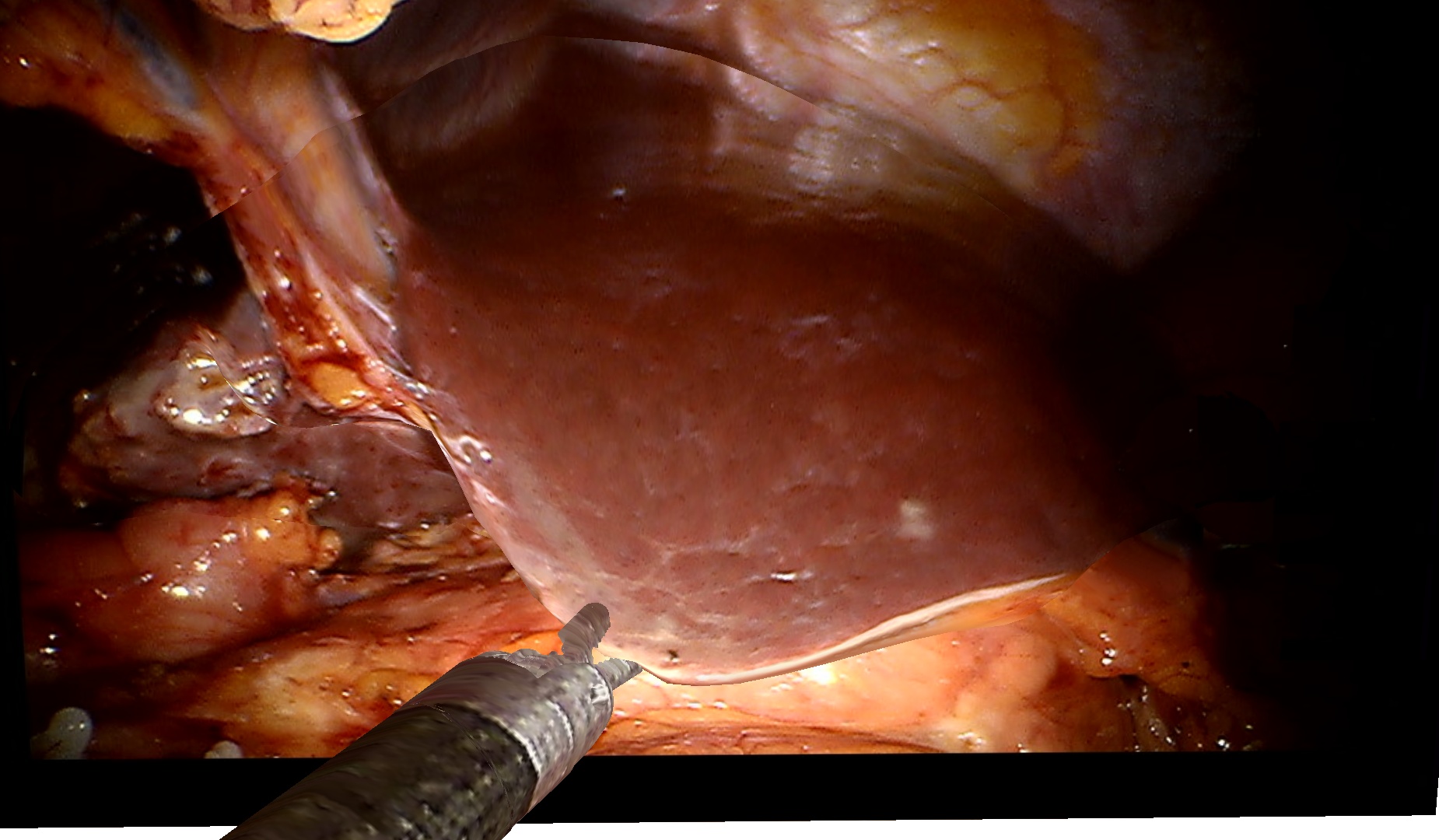}} \\
          
    \subfloat{\includegraphics[width=0.195\linewidth]{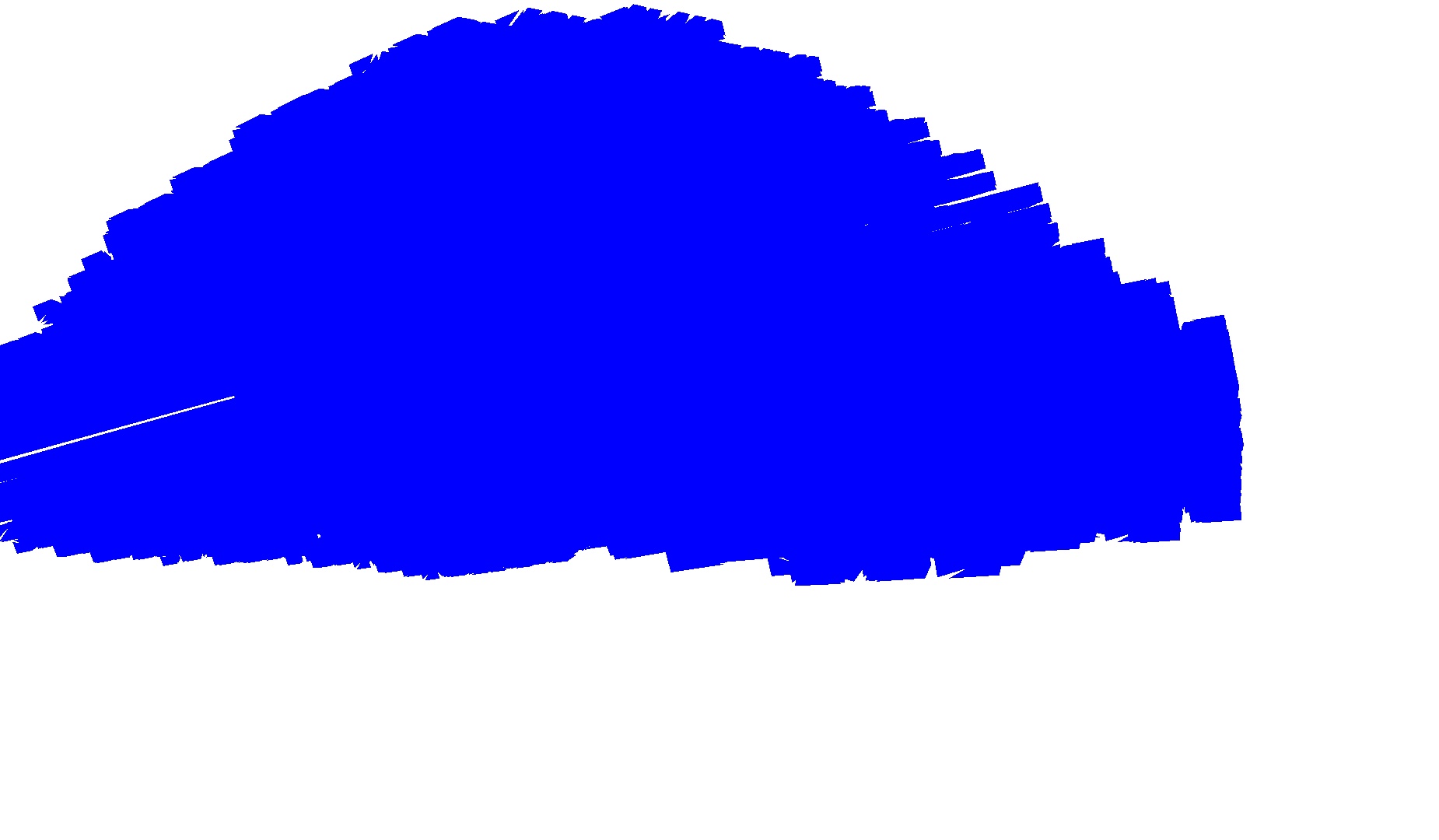}}
    \hfill
    \subfloat{\includegraphics[width=0.195\linewidth]{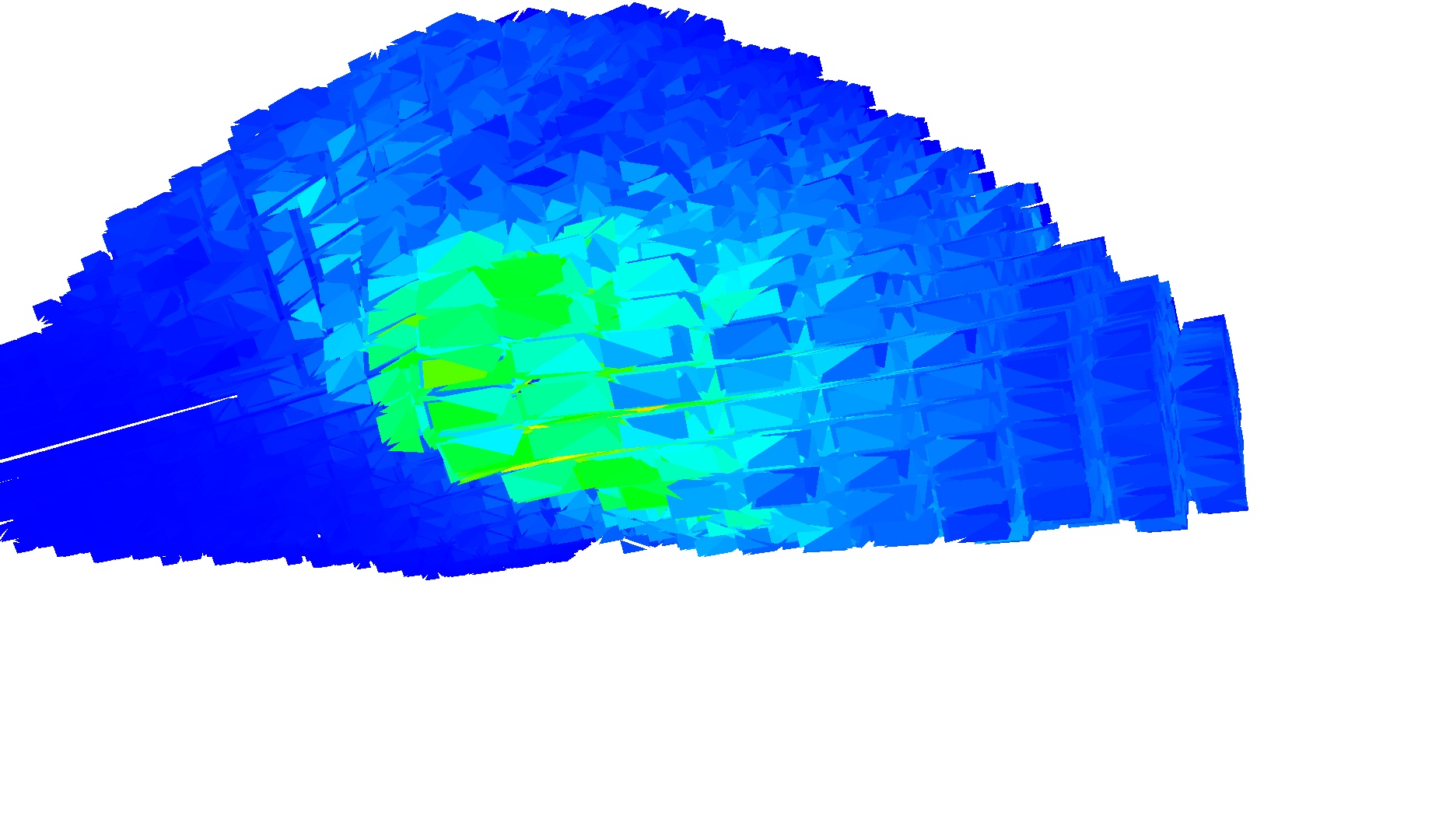}}
    \hfill
    \subfloat{\includegraphics[width=0.195\linewidth]{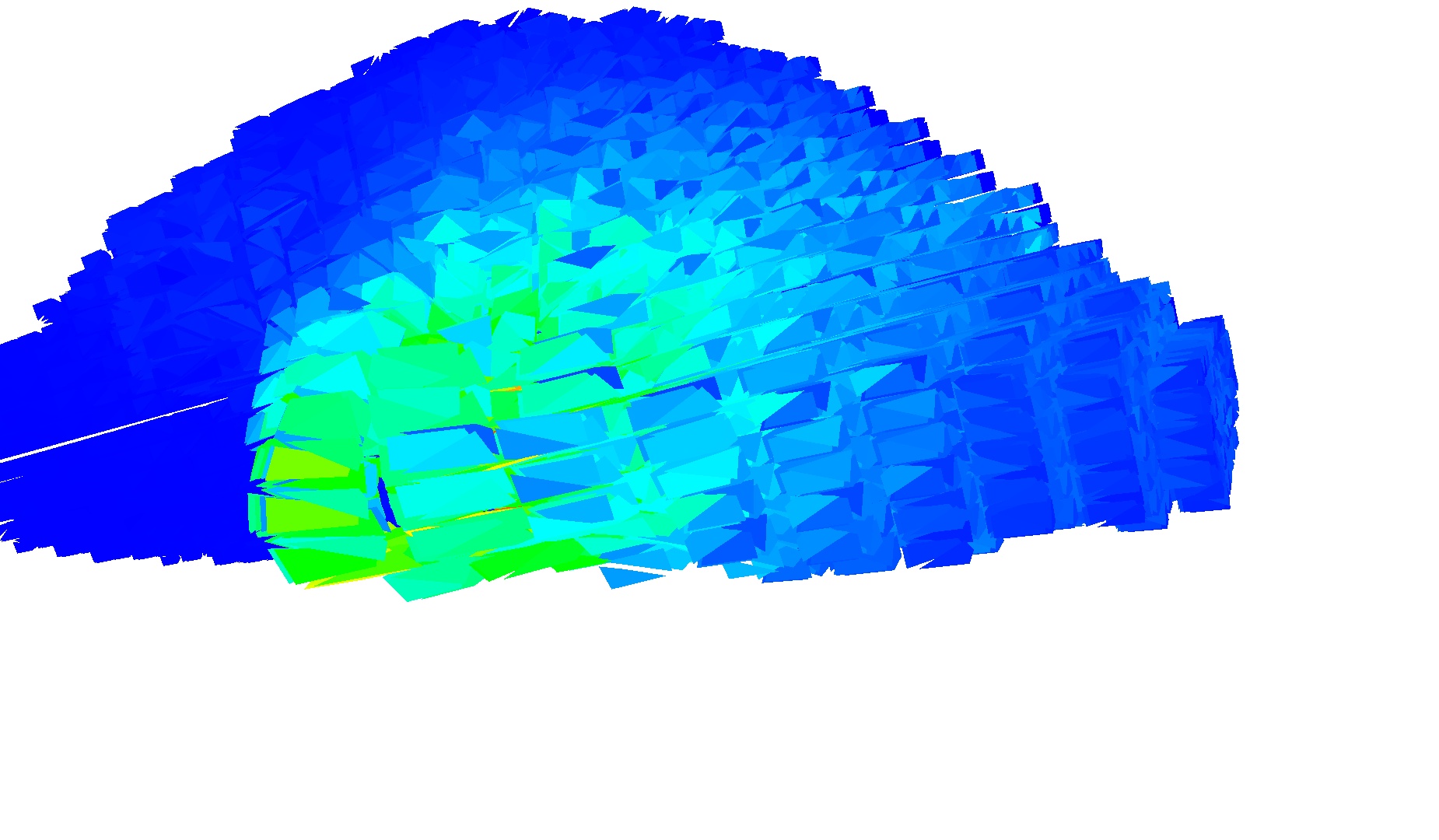}}
    \hfill
    \subfloat{\includegraphics[width=0.195\linewidth]{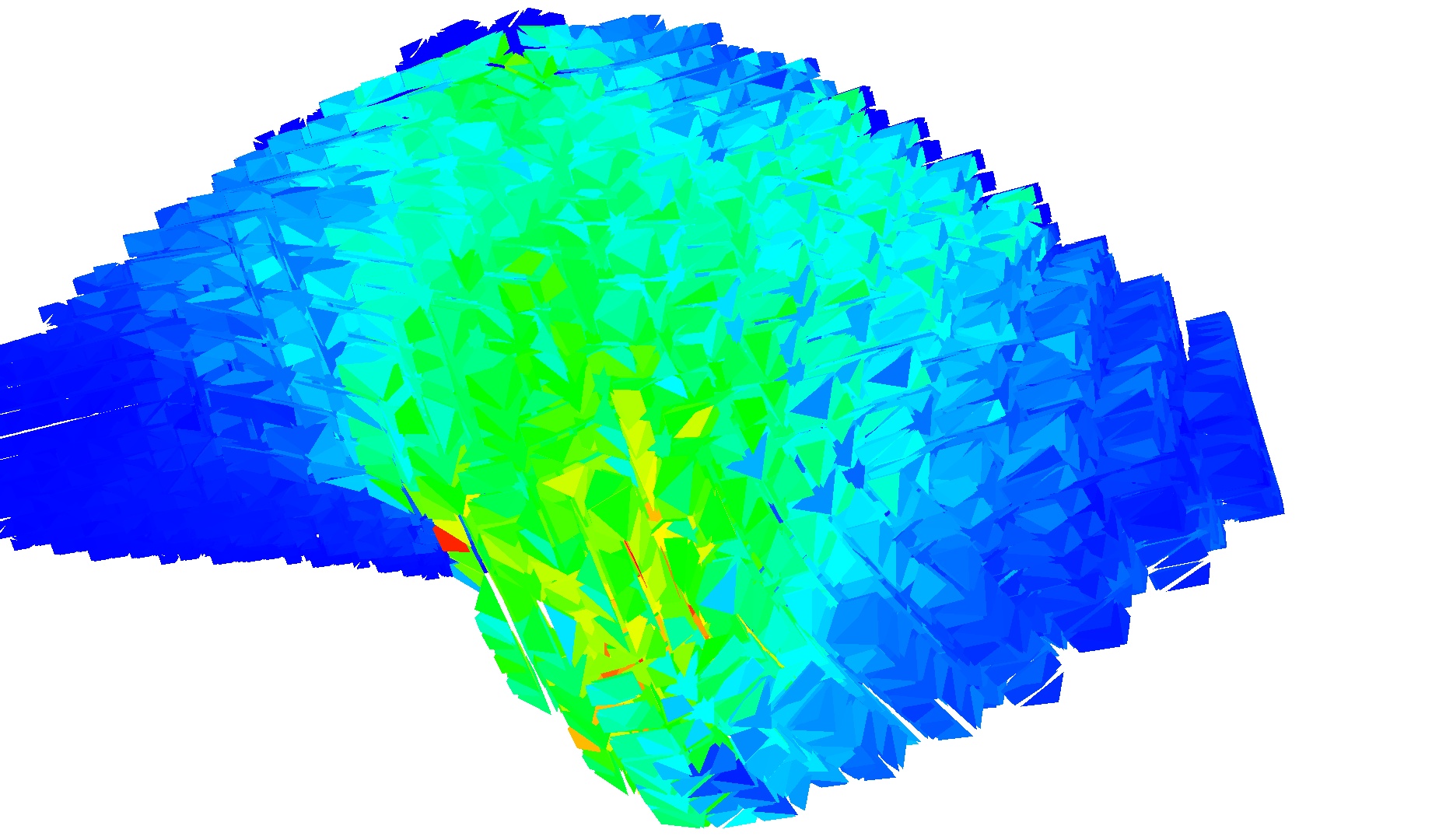}}
    \hfill
    \subfloat{\includegraphics[width=0.195\linewidth]{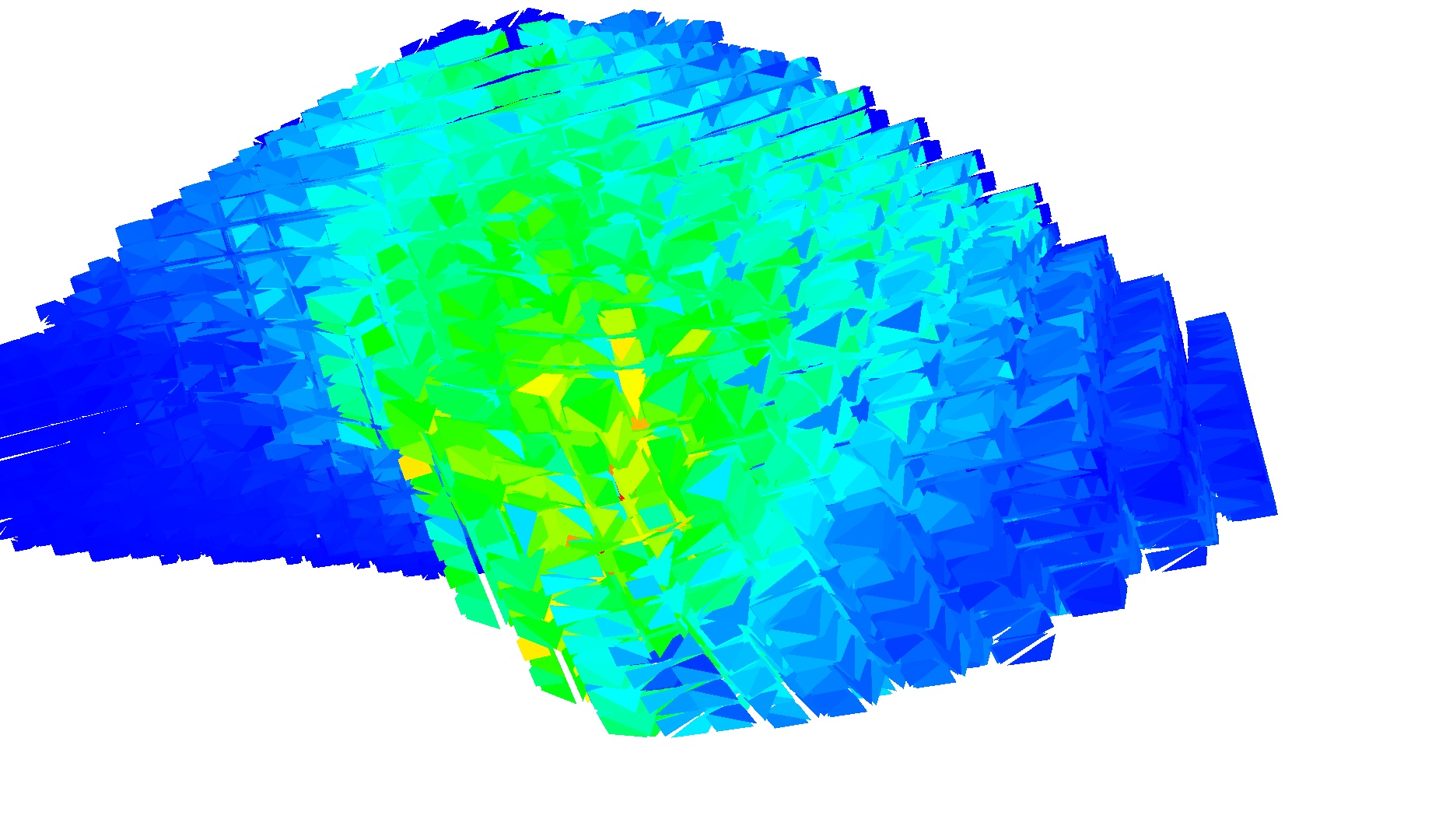}}
    \caption{Example sequence of a simulated DejaVu \cite{haouchine2017dejavu} scene: on top, the textured mesh, tool, and a realistic background, underneath the 3D model colorized based on stress.}
    \label{fig:sofa_full_scene}
\end{figure}

We generate five synthetic tool-tissue interactions on a human liver, using DejaVu simulation \cite{haouchine2017dejavu}. DejaVu provides realistic surgical scenes by transferring the appearance of real surgical images to physics-based simulations. The deformations are computed using the Finite Element Method in SOFA \cite{sofa}. Examples are shown in \Cref{fig:sofa_full_scene}.

For each sequence, we perform 45 simulation steps. We have an initial mesh and at each step in the simulation, we extract the image without the surgical tool, corresponding camera parameters, and the underlying mesh ground truth. We generate five different force profiles, yielding five simulated sequences with 45 images, camera parameters, and mesh ground truths each. To cover a wide range of interactions, we choose a different predominant manipulation in each sequence: pushing into the liver (SimIn), pulling to the left (SimLeft), pulling down (SimDown), dragging in a circular motion (SimCircular), and pulling up (SimUp).

\begin{figure}[!ht]
    \centering
    \includegraphics[width=0.95\linewidth]{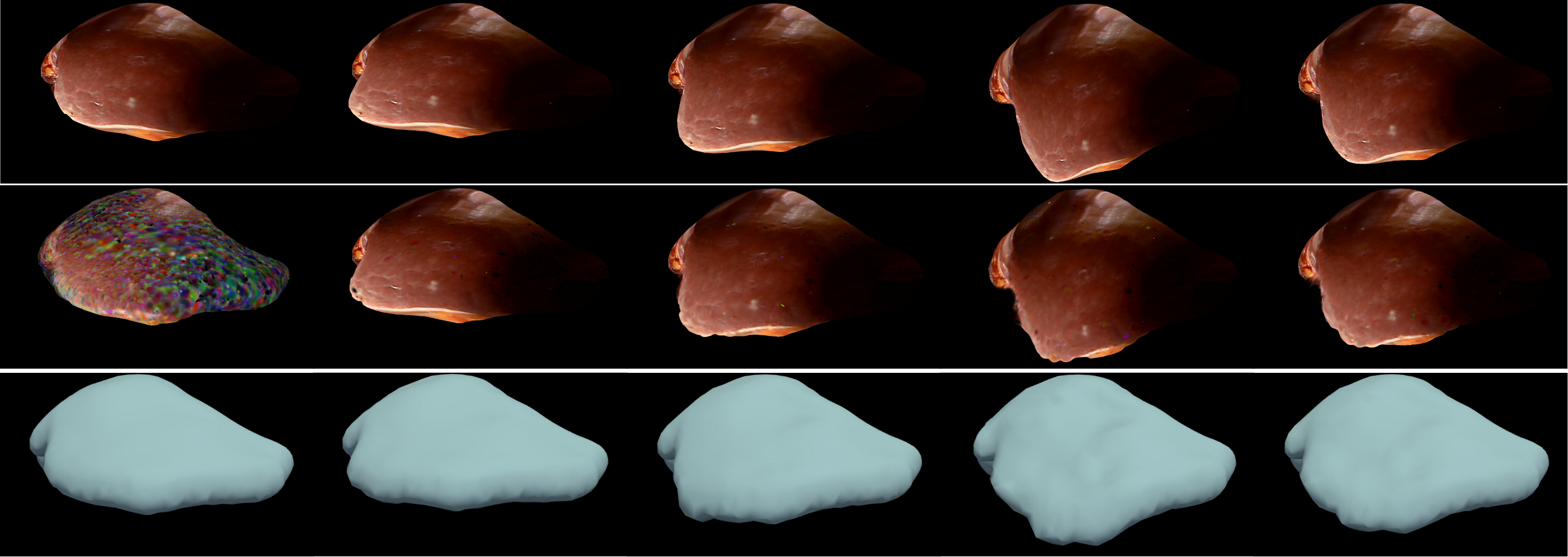}
    \caption{Exemplary results on a simulated DejaVu \cite{haouchine2017dejavu} sequence. Rows top to bottom: input image with surgical tool and background image removed, rendered image from Gaussian Splatting, and deformed mesh. The first column shows the first step of the sequence, where some of the Gaussians still have random colors. See video in the supplementary material.}
    \label{fig:examples_synthetic}
\end{figure}

We run our method on the five sequences and in each simulation step, we compute the Euclidean distance between each vertex in our deformed mesh and the corresponding vertex in the ground truth mesh. For each simulated sequence, we compute the average Euclidean distance, its standard deviation, and the maximum Euclidean distance.
As reported in \Cref{tab:dejavu_results}, we stay within a 5~mm error margin and observe only slight deviations in accuracy across the different interactions, indicating that our method can handle in-plane (e.g., SimLeft) and out-of-plane deformations (e.g., SimIn).
A selection of tracked deformations on the DejaVu dataset is shown in \Cref{fig:examples_synthetic}.

\begin{table}
    \centering
    \caption{Quantitative evaluation on the DejaVu dataset with Euclidean distance and maximum error (in mm) on five simulated tool-tissue interaction sequences.}
    \label{tab:dejavu_results}
    \resizebox{0.95\textwidth}{!}
    {
    \begin{tabular}{l|ccccc}
        \toprule
        Metric & SimIn & SimDown & SimLeft & SimCircular & SimUp \\
        \midrule
        \textbf{Euclidean Distance (mm)} & 0.11$\pm$0.19 & 0.14$\pm$0.27 & 0.09$\pm$0.16 & 0.10$\pm$0.19 & 0.12$\pm$0.24 \\
        \textbf{Max Euclidean Distance (mm)} & 3.17 & 4.29 & 2.78 & 4.15 & 2.84 \\
        \bottomrule
    \end{tabular}
    }
\end{table}

\subsubsection{Qualitative Evaluation on Clinical Data}
To demonstrate the effectiveness of our method on clinical data, we use a dataset of two visceral pig surgeries with a preoperative post-insufflation CT, initial rigid registration, a tracked laparoscope, and corresponding 2D monocular RGB videos. Surgical tools were masked out using SurgicalSAM \cite{yue2024surgicalsam}. The first sequence from the first surgery shows a tool-tissue interaction, whereas the second one shows a breathing motion without external manipulation. Because there is no deformed ground truth, as measuring intraoperative deformation would require intraoperative scanning, we demonstrate qualitative results.

\begin{figure}
    \centering
    \includegraphics[width=0.8\linewidth]{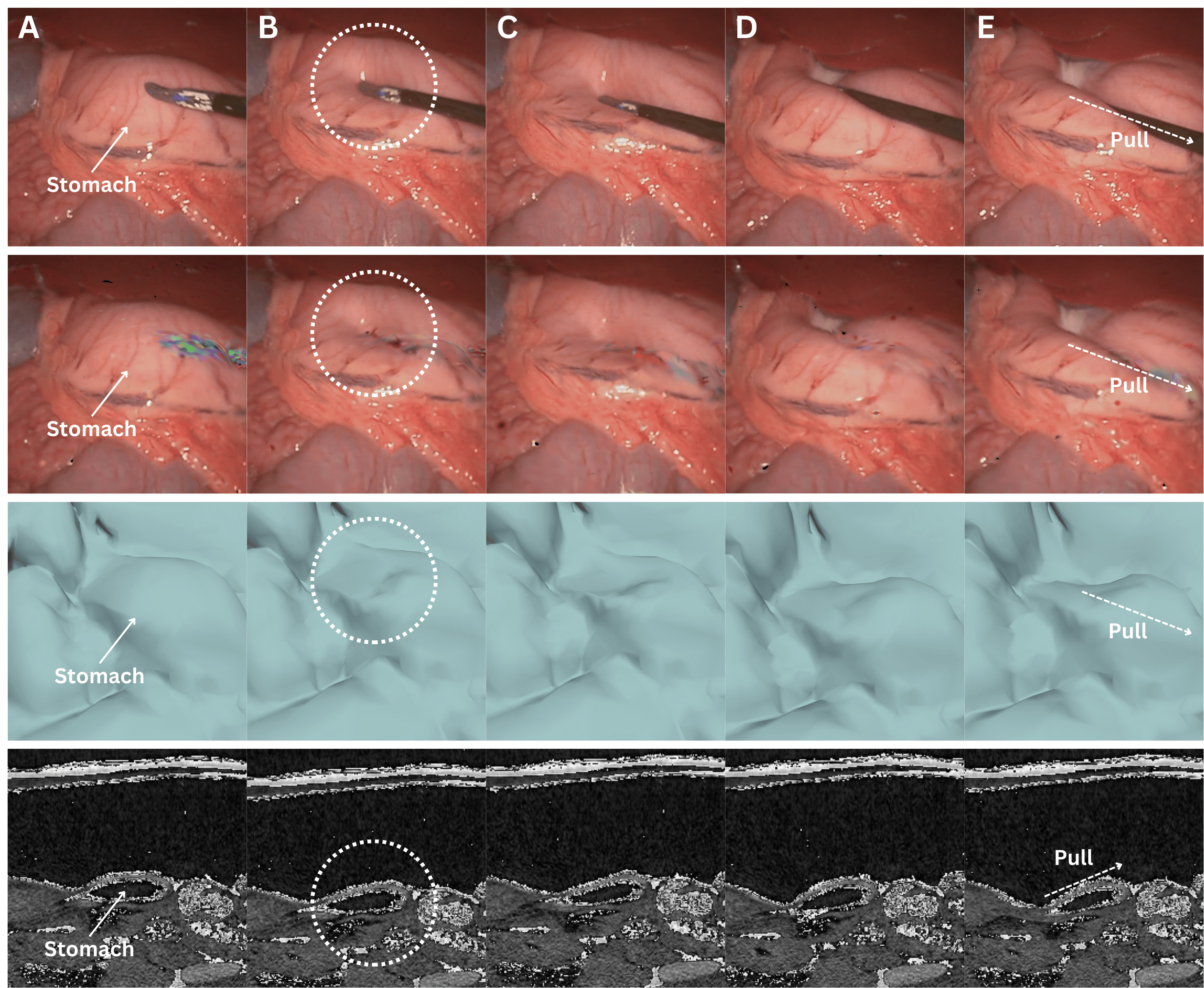}
    \caption{Examples from a tool-tissue interaction sequence. Rows from top to bottom: input images acquired by the laparoscope, images rendered with our method, deforming mesh, and deforming CT. Arrows in A point to the deforming anatomical structure, circles in B highlight the area of deformation, and arrows in E visualize the pull direction. See video in the supplementary material.}
    \label{fig:examples_clinical}
\end{figure}

\Cref{fig:examples_clinical} shows results on the tool-tissue interaction. The surgical tool presses down on the stomach, leading to the deformations tracked in the deformed mesh and CT (bottom rows). Although the difference in pressure applied between B and C appears marginal, we successfully track the increasing deformation, the difference being visible in both mesh and CT. For D and E, the tool is repositioned, now pulling the stomach back. Note how this backward pull is visible in the respective deformed CTs by a deformation to the \emph{right}, since the laparoscope captures the scene roughly in the coronal plane, whereas the CT slice is shown in the sagittal plane.

Since we rely on an initial registration, we are also subject to registration errors. There is a slight misalignment between the structure being manipulated in the input images (stomach) and the deforming mesh. Also note that we initialize our mesh from a post-insufflation CT, ensuring that the initial mesh and the intraoperative state early in surgery are still very similar. Regarding the deformation of the CT, we can only capture surface deformations by deforming our mesh. A more realistic deformation of deeper layers would require biomechanical modeling.
Apart from those caveats, the method does not work in real time yet. Given the strides being made in the 3D computer vision community around 4DGS, we are confident that this problem will be solved soon. Despite those limitations, the deformations of the stomach in the CT are sensible and correspond to the tool-tissue interaction as confirmed by a board-certified surgeon.

\begin{figure}
    \centering
    \includegraphics[width=1\linewidth]{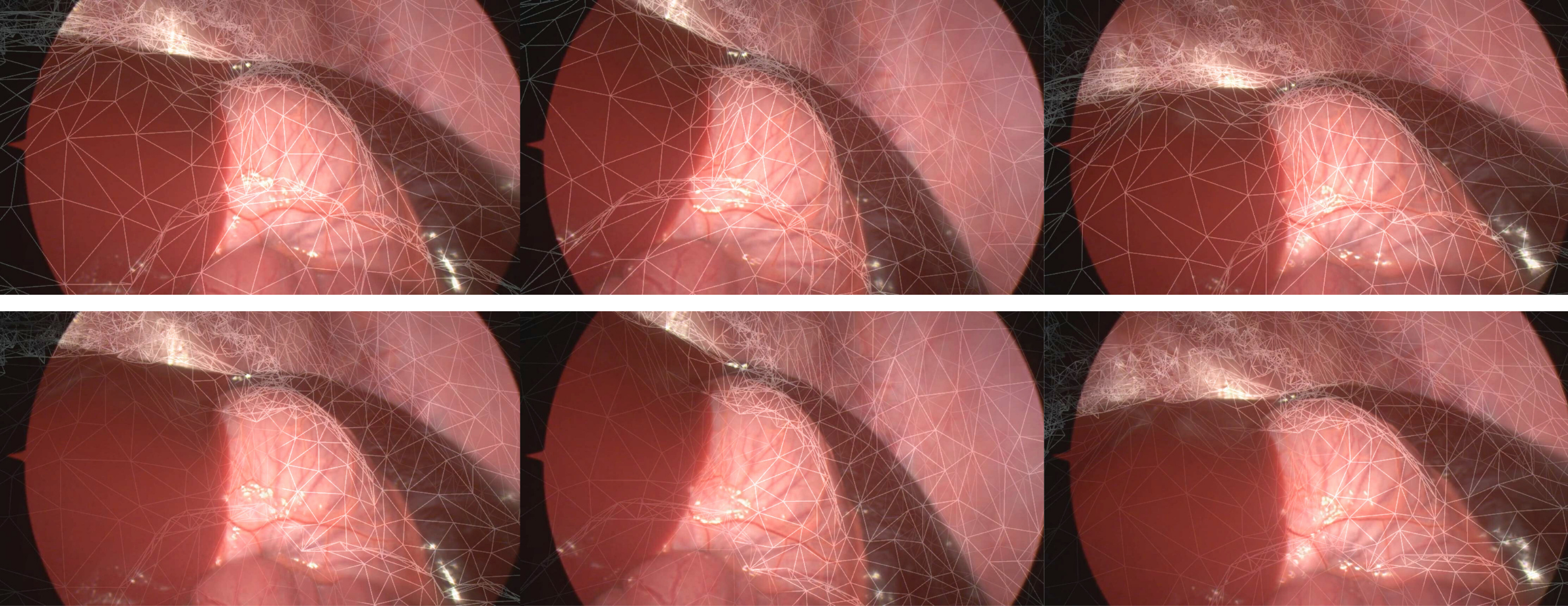}
    \caption{Top row: overlays based on rigid registration. Bottom row: deformed overlays with our method. Deformations are subtle (see video).}
    \label{fig:overlay}
\end{figure}

Coming from our clinical motivation, we also demonstrate potential dynamic overlays in \Cref{fig:overlay} for the sequence of the second surgery, displaying breathing motion. The overlays improve over the initial rigid registration, demonstrating the method's ability to pick up on subtle deformations.

\section{Conclusion}
In this paper, we have demonstrated for the first time that a direct coupling between CT and 4D Gaussian Splatting is possible to deform preoperative volumetric patient data during surgery. By blending the concepts of 4D Gaussian Splatting and deforming 3D meshes, we have shown that a mesh of the abdominal cavity can serve as an effective intermediary to capture changes observed through the laparoscope and relay them directly to the CT. We have evaluated our method on synthetic and clinical data, demonstrating that its accuracy meets clinical needs on simulated data and performs well under real conditions in visceral pig surgeries. Strikingly, this is possible on 2D RGB data even without the need for stereo laparoscopes or any depth information.
\begin{credits}

\end{credits}


\bibliographystyle{splncs04}
\bibliography{mybib}

\end{document}